\documentclass[10pt,twocolumn,letterpaper]{article}
\pdfoutput=1
\usepackage{iccv}
\usepackage{times}
\usepackage{epsfig}
\usepackage{graphicx}
\usepackage{amsmath}
\usepackage{amssymb}

\usepackage[boxed,titlenumbered,linesnumbered,algoruled]{algorithm2e}

\usepackage[pagebackref=true,breaklinks=true,letterpaper=true,colorlinks,bookmarks=false]{hyperref}

\iccvfinalcopy 

\ificcvfinal\fi
\begin{document}

\title{HiCoRe: Visual Hierarchical Context-Reasoning}

\author{Pedro H. Bugatti$^{1,2,}{\thanks{Equal contribution}}$ , Priscila T. M. Saito$^{1,2,*}$, Larry S. Davis$^{2}$\\
$^{1}$Federal University of Technology - Parana \hspace{0.6cm} $^{2}$University of Maryland, College Park \\
{\tt\small \{pbugatti, psaito\}@utfpr.edu.br \hspace{0.6cm}  lsd@umiacs.umd.edu}
}


\maketitle

\begin{abstract}
Reasoning about images/objects and their hierarchical interactions is a key concept for the next generation of computer vision approaches. Here we present a new framework to deal with it through a visual hierarchical context-based reasoning. Current reasoning methods use the fine-grained labels from images' objects and their interactions to predict labels to new objects. Our framework  modifies this current information flow. It goes beyond and is independent of the fine-grained labels from the objects to define the image context. It takes into account the hierarchical interactions between different abstraction levels (i.e. taxonomy) of information in the images and their bounding-boxes. Besides these connections, it considers their intrinsic characteristics. To do so, we build and apply graphs to graph convolution networks with convolutional neural networks. We show a strong effectiveness over widely used convolutional neural networks, reaching a gain $3$ times greater on well-known image datasets. We evaluate the capability and the behavior of our framework under different scenarios, considering distinct (superclass, subclass and hierarchical) granularity levels. We also explore attention mechanisms through graph attention networks and pre-processing methods considering dimensionality expansion and/or reduction of the features' representations. Further analyses are performed comparing supervised and semi-supervised approaches.
\end{abstract}

\section{Introduction}
\label{sec:introduction}

Deep neural networks (e.g. Convolutional Neural Networks - CNNs) are effective in several computer vision tasks such as image segmentation \cite{Olaf2015, Lin2016, Yao2018, Ruirui2019}, classification \cite{He2016, Sandler2018, Pu2019, Bu2019}, object detection \cite{Liu2016, Ren2017, He2017, Zeng2018}, among others. However, a crucial problem remains: how to deal with context?  

In real environment global context definition is common and computer vision systems must handle it.
However, under such scenarios, traditional end-to-end CNN architectures collapse. This occurs because CNNs and literature works ignore the semantic relations among objects to define the image context. Hence, we believe that to answer the first question we need to solve another one: how to grasp and describe the semantic relations among objects to define the context (global class) of the image?

Although CNNs are considered the ``holy grail'' of image recognition, they are incapable to answer this question. CNNs fail because they are based on feature maps and cover the eyes to the relations between objects. Our insight is that, semantics between objects can be obtained through their visual and spatial interactions. To reach this semantic reasoning our method fuses Graph Convolutional Networks (GCNs) \cite{Defferrard2016, Zhou2018, Zhang2018, Battaglia2018} and CNNs. We focus on the proposal of a visual hierarchical context-reasoning deep learning architecture capable of defining the global context of an image in different granularities.

\begin{figure*}[h!]
\centering
\includegraphics[width=16cm]{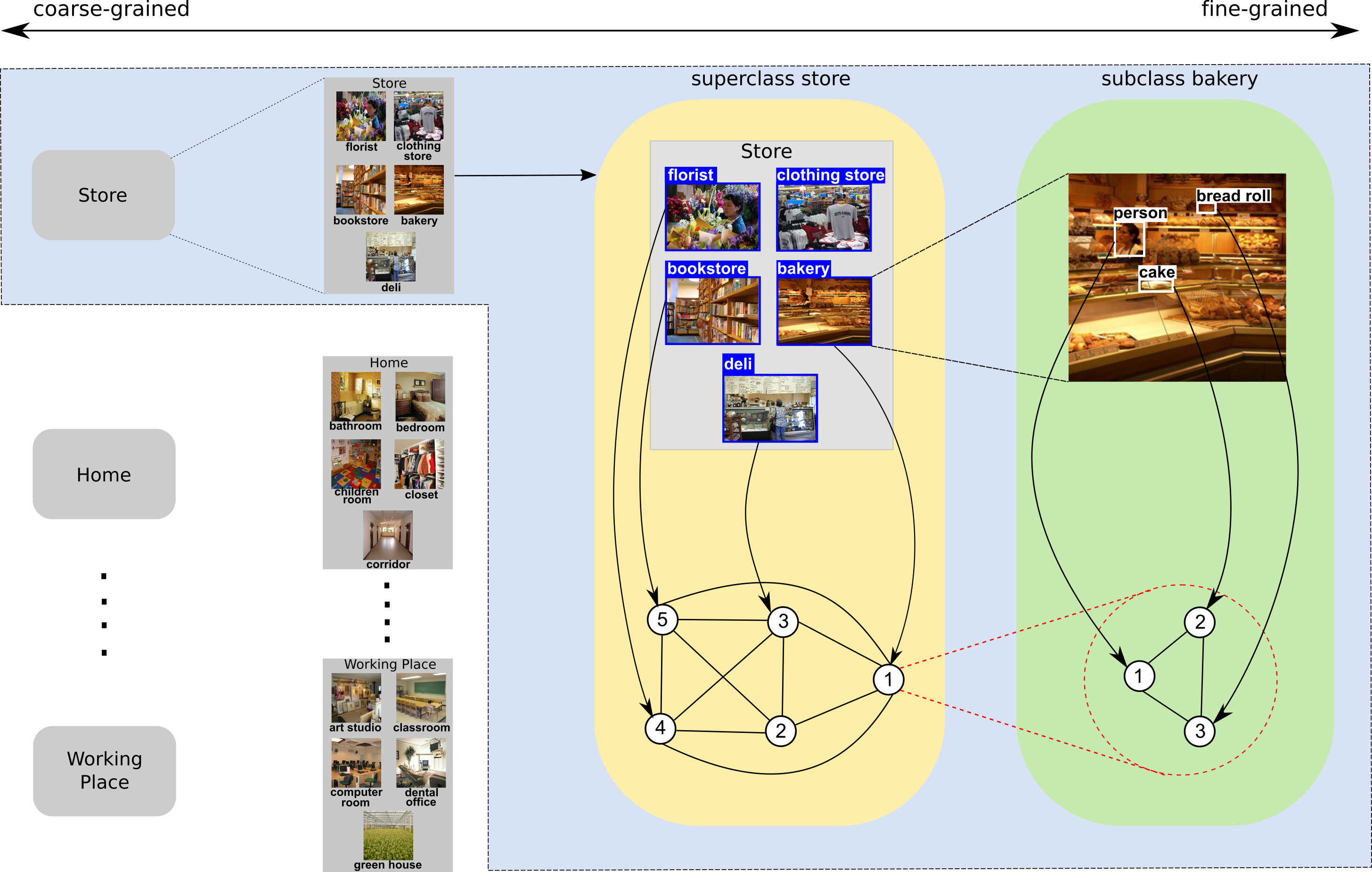}
\caption{Current computer vision systems are incapable to reach a correct reasoning over and above convolutions. To fill this gap, we proposed a visual hierarchical context-reasoning deep learning architecture capable of adding visual and contextual information to the images' objects. Here we illustrate an example of our proposed hierarchical interaction, considering different context granularities, i.e. objects, images, and superclasses. Images are from the dataset available in \cite{Quantoni2009}.}
\label{fig:example}
\end{figure*}

Figure \ref{fig:example} illustrates a problem with different context granularities (coarse to fine-grained). For instance, a coarse-grained information is related to superclasses, such as ``store'', ``home'', ``working place'', etc. Each superclass is composed of different subclasses of images (medium-grained level). Considering the superclass ``store'' it holds the subclasses ``florist'', ``bookstore'', ``clothing store'', ``bakery'' and ``deli''. The fine-grained level considers the bounding-boxes (e.g. ``cake'', ``person'' and ``bread roll'' belonging to the subclass ``bakery''). To predict the global context of the image our framework can build different graph representations for each granularity level. Each graph node encodes intrinsic information from the image or a bounding-box.

Our approach uses the fine and coarse-grained levels to build, respectively, inner (bounding-boxes of an image) and outer (superclass node linked to its subclasses nodes) interactions between nodes from the graph. Using this structure we can better grasp and describe the context in a hierarchical way. This occurs, because we capture how images and their objects (bounding-boxes) interact with each other. Our framework can cope with different levels of granularity.


Despite efforts regarding graph convolutional network to capture object interactions in the image, to the best of our knowledge, previous literature differs from our proposal.

The approaches \cite{Feifei2017, Yang2018, Li2018, Li2017} focus on the scene graph generation paradigm. These papers proposed different models to detect objects, their categories and relations (i.e. predicates of pairwise relationships and object categories). Our approach differs from this paradigm. It is not based on object detection, its category or relations. Our goal is to define the global context of the image which is completely different from these works that want to generate a scene graph (inference of predicate and object). Our approach does not need to infer predicates or object categories to define the global context. The only ground truth we need are the proposals' bounding-boxes and the global class of the image. Moreover, to diminish the computational complexity, we consider a complete graph between the objects of an image, not depending on a pruning pre-processing as \cite{Yang2018, Li2018}. In addition, we did not use extra supervision through RNNs feedback like \cite{Feifei2017, Yang2018}.


Works like \cite{Gupta2019, Lin2018} present some similarities with ours. However, they present problems. In \cite{Gupta2019} the authors overlook the hierarchical context idea (granularities). Besides, they use a reinforcement learning paradigm worsening the complexity of the model. Their approach also needs to know the objects' classses and requires an a priori knowledge graph. The authors use classification scores (from ResNet-50) and word embedding to represent each node.

On the other hand, our approach is independent of a prior knowledge (i.e. graph), and the objects' classes are unnecessary. Besides, it does not require previous classification score or embedding methods. We use simple visual and spatial features presenting a lower computational cost. We also consider a single type of graph node to be capable of generalizing to various problems. 

In \cite{Lin2018}, the authors consider different types of nodes (i.e. relationship and object nodes). To generate a relationship node they crop three regions from the image. The first region covers the union of two objects and the other two contain each object. Then, they concatenate the feature vectors from these three regions. It is a biased approach because the first region implies that the objects are already connected. This leads to an unfair comparison as it introduces more supervision and additional information. It was also used two architectures to extract the visual features (ResNet-101 and VGG-16), resulting in a higher computational cost.

In addition, the authors in \cite{Lin2018} use their own detected objects. They argued that the previous object detection task is necessary because no dataset meets the conditions to execute their approach. However, they did not provide the detected objects, which can cause a biased evaluation. It hinders the replication of their experiments, and a fair comparison. Differently, our method can use any dataset that provides object proposals and the image class. We do not even need the object class, reducing the supervision. These properties are important to achieve experiments that can be reproduced by any researcher. We also evaluated our method under a higher number of classes and more challenging scenarios (hard relationship levels). Our approach shows strong performance over state-of-the-art convolutional neural networks.


\textbf{Contributions:} In summary, the contributions of this work are threefold. First, we developed a new visual Hierarchical Context-Reasoning (HiCoRe) framework that can predict the global context of images regardless of knowing the labels of objects from an image. For instance, it can cope with problems where the labels from the bounding-boxes are partially or totally missing. Other innovation is that our framework is capable of defining the global context through a hierarchical scheme, which considers different (superclass, subclass and hierarchical) granularity levels. HiCoRe takes into account not only the interactions between the bounding-boxes and their intrinsic characteristics, but also the connections between the images from a same context. Second, we analyzed the capability and the behavior of our framework under different scenarios through reduction and expansion methods applied on object/image descriptions. To consider edge relevance we also applied a recent weighting graph technique based on graph attention networks. Third, we explored our framework considering supervised and semi-supervised approaches.

\section{Background and Related Works}
\label{sec:background}

Our framework is inspired in previous works \cite{Kipf2017, Skiena2018} that try to grasp the interactions and the structure of a graph's nodes through neural networks. Work of \cite{Skiena2018} proposes to learn low dimensional embeddings of a graph's nodes. To do so, they decompose a graph in several levels (coarse to fine-grained) and preserve the graph structural features. In \cite{Kipf2017} the authors proposed a scalable learning method to convolutional neural networks on graphs called Graph Convolutional Network (GCN). The work consists of using multiple graph convolution layers that promote a neighborhood aggregation of information. After $L$ layers a given node fuses the information from its neighbors that are $L$-hops of distance in the graph. In other words, GCN modifies the feature vector, performing a kind of feature propagation, by aggregating more and more information at each hidden layer (i.e. $l$ layer). 

Considering an undirected graph defined as $G = (V, A)$, where $V = \{v_1, v_2, ..., v_N\}$ is a set of $N$ vertices (nodes), and $A \in \mathbb{R}^{N \times N}$ is a symmetric adjacency matrix where its element $A_{ij}$ is $1$ when there is an edge from node $i$ to node $j$, and $0$ otherwise. Each node $v_i$ is represented by a feature vector $x_i \in \mathbb{R}^d$, where $d$ is the space dimensionality. Therefore, we have a feature matrix $X \in \mathbb{R}^{N \times d}$ that stacks the feature vectors (i.e. $X = [x_1, x_2, ..., x_N]$).

\begin{figure*}[h!]
    \centering
    \includegraphics[width=17.4cm]{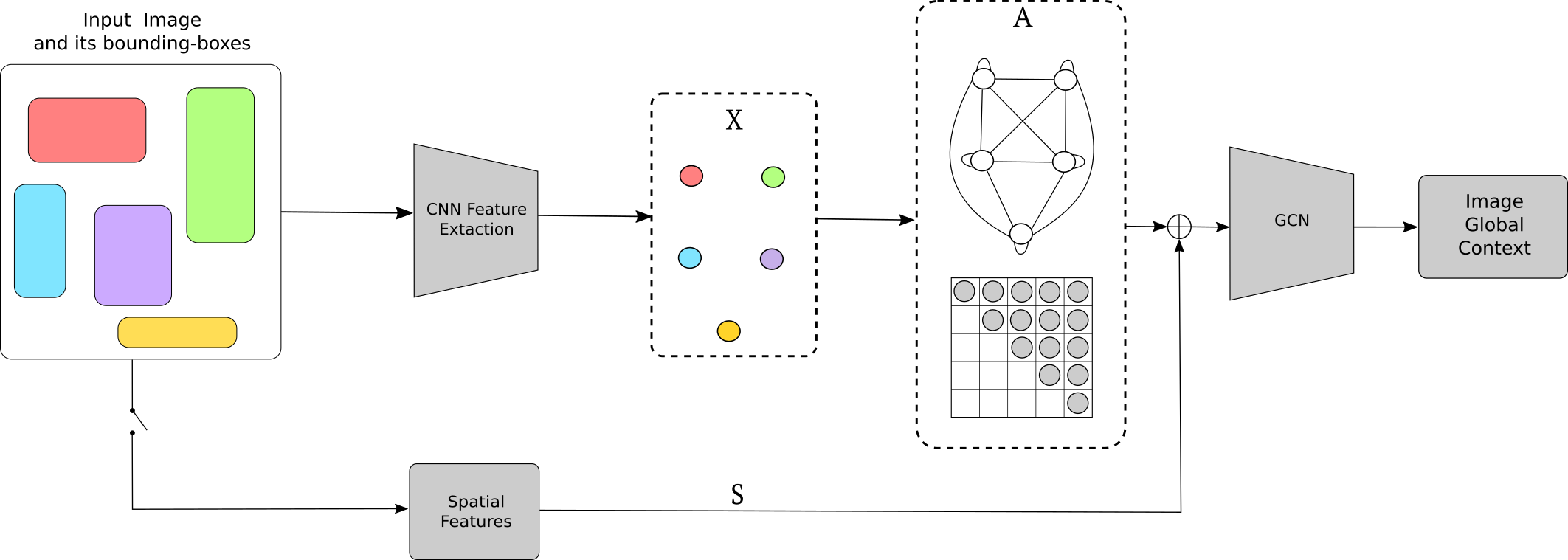}
    \caption{Pipeline of our proposed visual Hierarchical Context-based Reasoning (HiCoRe) framework. HiCoRe deals with different granularity levels. Here we see the representation of the object-level granularity (subclasses). The bounding-boxes from the images are described by a visual encoder (pre-trained CNN). For each image, its bounding-boxes will be nodes in a complete graph (i.e. they interact with each other). An adjacency matrix is built and encodes the relationships between the bounding-boxes. Then, a GCN joins the (visual and/or spatial) features from each node with its interactions, propagating the information and resulting in our global context (without requiring knowledge of the bounding-boxes' labels).}
    \label{fig:pipeline}
\end{figure*}

At each GCN layer the features of each node is averaged with its respective neighboring feature vectors, leading to a feature propagation process. In \cite{Kipf2017} the authors simplify this feature propagation by using only the first-order neighbors (1-step neighborhood around each node). The propagation rule of a GCN is formally defined by Equation \ref{eq:ruleGCN}. The initial nodes' representations come from the original feature matrix ($H^{(0)} = X$).

\begin{equation}
   H^{(l+1)} = \sigma(\widehat{A}H^{(l)}W^{(l)})
   \label{eq:ruleGCN}
\end{equation}

\noindent where $l$ denotes the index of the graph convolution layer; $\sigma$ is the activation function (e.g. ReLU) for all but the output layer; $W$ is a trainable weight matrix for layer $l$; $\widehat{A}$ is a normalization of $A$ with added self-loops, defined as:

\begin{equation}
    \widehat{A} = \widetilde{D}^{-\frac{1}{2}} \widetilde{A}  \widetilde{D}^{-\frac{1}{2}}, \ \ \ \widetilde{A} = A + I_N
\end{equation}

\noindent here $I_N$ is the identity matrix and $\widetilde{D} = diag(\sum_{j}\widetilde{A}_{ij})$ is the degree matrix of $\widetilde{A}$.

Considering the output layer (i.e. node's classification), generally, it is used a softmax function. In this case, $H^{(L)}$ denotes the probability of a given node belongs to class $c$, where each node from a given class is represented by a $C$-dimensional one-hot vector $y_i \in \{0,1\}^C$.

Although successful, those GCNs proposed in \cite{Kipf2017} were evaluated in citation networks for document classification. They neglect the computer vision area. Different works in the literature propose the use of GCNs and its variants in computer vision \cite{Gupta2018, Marino2017, Chen2017}. However, these works are based on object recognition tasks. We are not interested in this kind of tasks. Our framework's goal is to define the global meaning (i.e. class) of a given image through its objects.

Works like \cite{Gupta2018, Chen2017} focus on region classification, aiming to assign labels to objects' regions specified by bounding-boxes. Besides, they require the ground-truth locations of such regions for both training and testing. In contrast, our approach does not need to know the class of each object in an image for training and testing our framework to reach the context-reasoning. This leads to a great advantage because we can bypass the detection and classification of objects, which is a labor intensive process. Moreover, unlike previous works \cite{Gupta2018, Marino2017, Chen2013}, HiCoRe is independent of an a priori knowledge graph to accomplish our context recognition, because our approach builds a complete graph.

We believe that to reach a successful visual reasoning, we should consider not only local and global information, but also the interaction between objects and their different layers of abstraction (granularities). By not considering all this information in a given hierarchical image structure it can lead to a misleading context recognition for tasks that involve regions (e.g. a group of images or bounding boxes that compose an abstract concept).

\section{Hierarchical Context-Reasoning Framework}
\label{sec:proposed-framework}

In this section we detail our visual Hierarchical Context-Reasoning (HiCoRe) framework. It describes the semantic relationships between bounding-boxes and define the image context through visual features and a graph structure. Besides the visual features, obtained from a pre-trained CNN, HiCoRe presents a main core.

\subsection{Context-Reasoning Core} 

The cornerstone of our framework is the context reasoning core (regions-images interaction module). It enables to model the semantic relationships of objects from an image and between the images themselves.

Our context-reasoning core deals with different granularity levels. Figure \ref{fig:pipeline} illustrates the representation of the object-level granularity (subclasses). The bounding-boxes from the images are described by a pre-trained CNN. 

Each bounding-box from an input image will be a node in a graph (interact with each other). To do so, our main core builds a graph $G = (V, E)$, where $V$ and $E$ represent the node set and the edge set, respectively. Next, it builds a complete graph to promote the information flow among all bounding-boxes, even those far from each other (and therefore not considered in a given receptive field). Then, the context-reasoning core creates an adjacency matrix to encode the relationships between the bounding-boxes. In the last phase, a GCN joins the features (visual and/or spatial) to propagate the information from each node.

Extensions can be applied to our framework, considering other types of features. We explored the inclusion of spatial features (e.g. the normalized translation between the bounding-boxes, the ratio of box sizes, among others). To better describe these spatial features we considered to expand them to a fine-grained representation through a Gaussian Mixture Model (GMM) discretization.

Considering this extension, we proposed a propagation rule to fuse visual and complementary features (e.g. spatial) into a GCN, and it is formally defined by Equation \ref{eq:ruleFramework}.

\begin{equation}
   H^{(l+1)} = \sigma(\widehat{A}H^{(l)}S^{(l)}W^{(l)}) 
   \label{eq:ruleFramework}
\end{equation}

\noindent where $\widehat{A}$ is the renormalized adjacency matrix with added self-loops; $H^{(0)}$ represents the input visual features for each node; and $S^{(0)}$ denotes the initial complementary features (i.e. describes each node or each pair of nodes in the graph).

From Equation \ref{eq:ruleFramework} it is clear to note that HiCoRe can fuse different types of features (e.g. visual and spatial) to describe a node from the graph. It can also consider separated visual and spatial features (i.e. w/o fusion) or other combinations. Moreover, our framework can be straightforwardly extended to accommodate other policies. HiCoRe is easily generalized to other types of information (e.g. obtained from graph structure, among others) to generate different propagation rules. Nevertheless, the scope of this paper is to present the basis to promote further extrapolations. 

The information related to the context is intrinsically detected by HiCoRe through the representations of the nodes joined with our graph construction. It considers not only each object in a given image and its interactions, but also the connection among images from the same context. This provides the hierarchical description and interaction, resulting in our global context.

\begin{figure*}[h!]
    \centering
    \includegraphics[width=17.0cm]{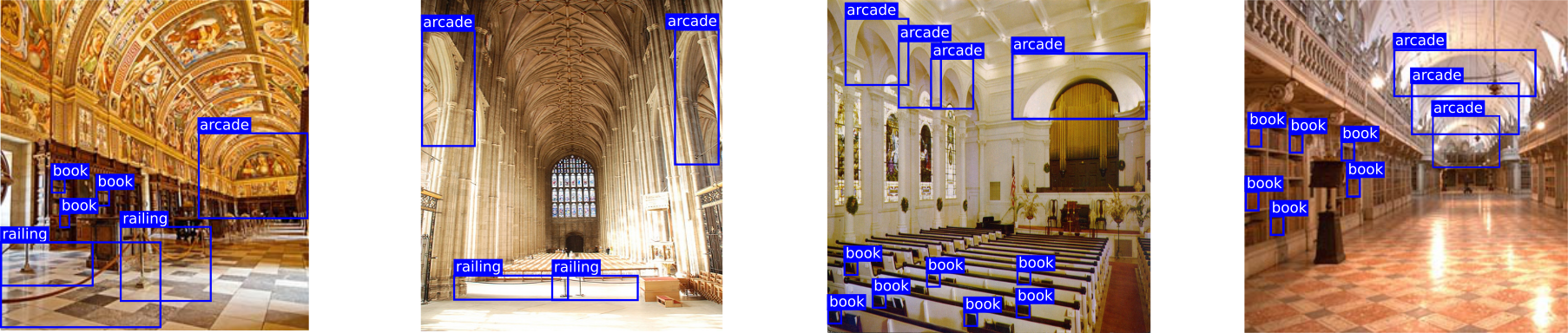}
    \caption{Examples of images in which the same bounding-boxes' concepts appear in different contexts. The two central images refer to the class ``church'' and the images of the left and right corners are related to the ``library'' class. Here we see that different semantic contexts (``church'' and ``library'') present the same classes for the bounding-boxes, such as: ``book'', ``railing'' and ``arcade''. This scenario brings ambiguity and compromises the classifications of traditional end-to-end CNN architectures. Images used in this example are from the public dataset provided in \cite{Quantoni2009}.}
    \label{fig:ambiguity}
\end{figure*}

To the best of our knowledge this is the first time that GCN are used to capture the global context (i.e. class) of an image through the interaction between its objects and connections between images from the same context. To train our network the labels of the objects are unnecessary. HiCoRe just need to use the global class of the image.

Our framework also works through a hierarchical interaction, considering different context granularities. For instance, once the bounding-boxes from an image (of a given subclass) are nodes in a graph structure, HiCoRe can consider each image in a given superclass as nodes in a supergraph that contains a subgraph of bounding-boxes nodes (see Figure~\ref{fig:example}).

In real environment ambiguities are common and computer vision systems must deal with it. However, under such scenarios traditional end-to-end CNN architectures generate compromised classifications. It occurs because the same bounding-boxes' concepts can appear in different contexts. For instance, considering two different images from divergent contexts: an image from subclass ``bar'' that has a context (superclass) defined as ``leisure'' has bounding-boxes from the classes ``chair'', ``table'', ``people''; and an image belonging to subclass ``classroom'' and associated with the context ``working place'' has bounding-boxes from the same classes. Using the bounding-boxes' classes to perform image classification will harm the final prediction, even in conjunction with other features from the image. This simple example testifies that state-of-the-art CNNs cannot cope with this issue, because different contexts can partially share the same objects and intrinsic features. Figure \ref{fig:ambiguity} illustrates this frequent problem. The two central images refer to the class ``church'' and the images from the corners are from the ``library'' class. It is clear to see that both distinct semantic contexts present a considerable overlap regarding their classes of bounding-boxes.

\subsection{Training}

Our framework is trained end-to-end and we consider as loss function the cross-entropy over all labeled samples. Algorithm \ref{alg:alg1-proposedApproach} details our training procedure. 

Although that was not the scope of this work, HiCoRe allows other loss functions because it can work with different feature representations. For instance, it is possible to propose a total loss composed by a specific loss applied to the visual features, other to the spatial features; and a final one to the different granularities when considering a hierarchical problem. Therefore, HiCoRe can be easily trained through back-propagation and with any optimizer (e.g. Stochastic Gradient Descent (SGD) \cite{Bottou2012}, etc).

To obtain a fair comparison between traditional state-of-the-art CNN architectures and our framework, we maintain the same loss (e.g. cross-entropy) to all approaches.

\begin{algorithm}[h!]
\caption{Training - HiCoRe}
\label{alg:alg1-proposedApproach}
\small
\SetKwInOut{Input}{input}
\SetKwInOut{Output}{output}
\SetKwInOut{Auxiliaries}{auxiliaries}
\Input{input images and their respective bounding-boxes}
\Output{hierarchical context-reasoning model}
$X \gets \emptyset$\;
\If{granularity = subclass}{
    $X \gets $ visual features of the bounding-boxes using CNNs\;
    $S \gets $ complementary (spatial) features (optional)\;
    \For{each image $i$ from dataset}{
        {\bf $V_i$} is composed of bounding-boxes\;
        Build a complete graph {\bf $G_i=(V_i,E_i)$}\;
    }
}
\ElseIf{granularity = superclass}{
    $X \gets $ visual features of the images using CNNs\;
    \For{each image $i$ from dataset}{
        {\bf $V_j$} is composed of images from a same context $j$\;
        Build a complete graph {\bf $G_j=(V_j,E_j)$}\;
    }
}
\uElse{ // if (granularity = hierarchical)\;
    $X \gets $ visual features of the images and the bounding-boxes using CNNs\;
    $S \gets $ complementary (spatial) features (optional)\;
    \For{each image $i$ from dataset}{
        {\bf $V_j$} is composed of images from a same context $j$ and their bounding-boxes\;
        Build a complete graph {\bf $G_j=(V_j,E_j)$}\;
    }
}

Compute the adjacency matrix {\bf $A$} from {\bf $G$}\;

\Repeat{{\bf \{$W$\}} has converged}{
Backpropagate and optimize {\bf \{$W$\}}\;
}

\Return{trained model}\;
\end{algorithm}

\section{Experiments and Results}
\label{sec:experiments}

We first validate the potential of our framework to learn a global-context on a dataset, called Unrel \cite{Peyre2017} (see Section \ref{subsec:unrel}). It comprises unusual relations between objects (i.e. not easily to be recognized). To analyze the behavior of our framework we also apply pre-processing methods like dimensionality expansion (e.g. Gaussian Mixture Models - GMM) \cite{Gader2018} and/or reduction (Principal Component Analysis - PCA) \cite{Zheng2017} of the features. Besides considering unweighted graphs, we perform analyses using weighted graphs through Graph Attention Networks (GATs) \cite{Velickovic2018, Boaz2018}.

We also corroborate the HiCoRe capability (of context-reasoning) through different levels of granularities (subclass, superclass), and considering their integration (hierarchical relations provided by our framework). To do so, we used the MIT67 image dataset \cite{Quantoni2009} (see Section \ref{subsec:mit}). Finally, we legitimize the ability of HiCoRe using the Visual Relationship Detection (VRD) \cite{Feifei2017} that presents a larger number of images and their respective bounding-boxes (for details see Section \ref{subsec:vrd}).

\subsection{Implementation Specifications} \label{subsec:impl-details}

To achieve the best results and a fair comparison, we performed a grid-search for the hyperparameters of the state-of-the-art CNNs, considering epochs ($2000$), learning rate ($0.001$, $0.005$, $0.01$, $0.05$), number of neurons ($16$, $32$, $64$, $128$, $256$) and dropout ($0.3$, $0.5$, $0.8$, $0.9$). We applied to each approach its best hyperparameters. For all image datasets we randomly split them, in a stratified way, generating their respective training ($80\%$ of images) and test sets ($20\%$).

To train HiCoRe, we make use of pre-trained models (VGG16 \cite{Zisserman2015}, VGG19 \cite{Zisserman2015} and ResNet50\cite{He2016}) on ImageNet \cite{Russakovsky2015} to extract the visual features from each image and/or bounding-box. We used such strategy to reach an independent task and, consequently, to learn a better visual description of the images, once lower levels of CNNs show these properties \cite{Razavian2014}. For instance, considering ResNet50 we obtained the deep features from the final conv5 layer after average pooling, which generates a feature vector with $2048$ positions. This same procedure was applied to VGG16 and VGG19 (using the fc7 layer) to obtain $512$-dimensional feature vectors. 

As an usual procedure in the literature, depending on the CNN architecture, the images and/or bounding-boxes need to be resized (enlarged or reduced), throughout the training and testing. HiCoRe applied these operations when necessary. Moreover, we used a two-layer GCN and we evaluated the accuracy of the predicted samples on the test sets. We chose this number of layers because in \cite{Kipf2017} the authors show that deeper models do not reach better accuracies, and can also lead to overfitting.

\subsection{Unrel Dataset} \label{subsec:unrel}

The Unrel dataset \cite{Peyre2017} consists of unusual relations. It contains images collected from the web with uncommon language triplet queries (e.g. ``person in cart'', ``dog ride bike'', among others). The images are annotated at box-level. In our experiments, we considered $822$ global images, $2,156$ bounding boxes and $59$ global classes.

We compared our proposed framework with the traditional end-to-end architectures (see Section \ref{subsec:impl-details} for implementation details). Different deep architectures, such as ResNet50, VGG-19 and VGG-16, were considered to analyze the most appropriate and to use it in the remaining experiments.

Our approach showed superior results compared to the traditional ones (see Table \ref{tab:results-unrel-architectures}). For instance, HiCoRe with ResNet50 reached an accuracy of almost $64\%$, while the traditional architecture reached about $35\%$. We observed this same behavior for VGG19 and VGG16, where our framework obtained accuracies of up to $62.31\%$ and $61.86\%$, respectively. Meanwhile, the traditional ones reached $42.13\%$ and $42.35\%$. In this sense, considering the best result (HiCoRe with ResNet50), our framework achieved an accuracy gain of $81.2\%$.

\begin{table}[h]
\caption{Comparison between HiCoRe and the traditional end-to-end ones, considering different deep architectures.}
\label{tab:results-unrel-architectures}
\centering
\begin{tabular}{lcc}
\hline
& {HiCoRe} & {traditional}\\\hline
{ResNet50} & $63.86$ & $35.25$\\
{VGG19} & $62.31$ & $42.13$\\
{VGG16} & $61.86$ & $42.35$\\
\hline
\end{tabular}
\end{table}

We also performed experiments to evaluate the impact of other configurations (relation representations) of objects in our approach. For instance, besides the visual features, we consider the spatial features \cite{Peyre2017} from each pair of bounding-boxes $o_p = [\alpha_p, \beta_p, \gamma_p, \delta_p]$ and $o_q = [\alpha_q, \beta_q, \gamma_q, \delta_q]$, where $(\alpha,\beta)$ are coordinates of the center of the box and $(\gamma,\delta)$ are the width and height of the box. Through the Equation \ref{eq:spatial}, we obtained a $6$-dimensional spatial feature vector.

\begin{center}
\begin{equation}
\small
  {r(o_p,o_q)}=\Big[\frac{\alpha_q - \alpha_p}{\sqrt{\gamma_p \delta_p}}, \frac{\beta_q - \beta_p}{\sqrt{\gamma_p \delta_p}}, \sqrt{\frac{\gamma_q \delta_q}{\gamma_p \delta_p}}, \frac{o_p \cap o_q}{o_p \cup o_q}, \frac{\gamma_p}{\delta_p}, \frac{\gamma_q}{\delta_q} \Big]
  \label{eq:spatial}
\end{equation}
\end{center}

The first two features represent the renormalized translation between the two boxes; the third is the ratio of box sizes; the fourth is the overlap between boxes, and the fifth and sixth encode the aspect ratio of each box, respectively. 

To obtain a well-suited representation, we perform the discretization of the feature vector 
into $k$ bins. For this, the spatial configurations $r(o_p , o_q)$ are generated by a mixture of $k$ Gaussians and the parameters of the Gaussian Mixture Model are fitted to the training pairs of boxes. We used as our spatial features the scores that represents the probability of assignment to each of the $k$ clusters. The spatial representation is a $400$-dimensional vector ($k = 400$).

From this fusion of (visual joined with spatial) features, we can note a slightly improvement (Table \ref{tab:results-unrel-spatial-features}). For instance, using ResNet50, we obtained $63.86\%$ of accuracy considering only the visual features, while with the fusion of the visual and spatial ones we achieved an accuracy of $64.10\%$.

\begin{table}[h]
\centering
\caption{Evaluation of the impact of different descriptions of objects.}
\label{tab:results-unrel-spatial-features}
\small
\begin{tabular}{lccc}
\hline
& {HiCoRe-} & {HiCoRe-} & {HiCoRe-}\\
& {ResNet50} & {VGG19} & {VGG16}\\\hline
visual & $63.86$ & $62.31$ & $61.86$\\
visual+spatial & $64.10$ & $61.86$ & $62.08$\\
\hline
\end{tabular}
\end{table}

In addition, as the extracted features have high dimensionality (mainly the visual features), we applied PCA to the visual and/or spatial features to analyze the impact caused by the reduction of the dimensions (Table \ref{tab:results-unrel-pca-features}). We used a PCA with $200$ components (i.e. reducing the dimensionality to $200$). From the obtained results, we can see that the dimensionality reduction is promising and can significantly improve the performance of our framework. 

Analyzing the results we can note that, even with an extreme reduction, we can maintain a good trade-off. For instance, regarding the visual features obtained from HiCoRe-ResNet50 (2048-D), we reduced the dimensionality $10.24$ times (i.e. from 2048-D to 200-D), and the accuracy was reduced by a factor of only $1.09$. This same behavior can be observed for all experiments. Moreover, as expected, when we applied PCA just on the spatial features the accuracy was less impacted. Considering visual and spatial features obtained from HiCoRe-VGG16 and with PCA applied just on spatial features, the accuracy presented a gain, reaching $63.19\%$ (see Table \ref{tab:results-unrel-pca-features}). On the other hand, with the same setting without PCA on spatial features the accuracy was $62.08\%$ (Table \ref{tab:results-unrel-spatial-features}).

\begin{table}[h]
\centering
\caption{Evaluation of the impact of dimensionality reduction.}
\label{tab:results-unrel-pca-features}
\small
\begin{tabular}{lccc}
\hline
& {HiCoRe-} & {HiCoRe-} & {HiCoRe-}\\
& {ResNet50} & {VGG19} & {VGG16}\\\hline
PCA(visual) & $58.31$ & $55.65$ & $54.77$\\
PCA(visual+spatial) & $58.54$ & $53.66$ & $54.10$\\
visual+PCA(spatial) & $63.41$ & $60.98$ & $63.19$\\
\hline
\end{tabular}
\end{table}

Once in our initial experiments we consider an unweighted graph we also conduct an analysis to verify the impact of considering weighted graphs. We performed experiments using Graph Attention Networks (GAT) \cite{Velickovic2018, Boaz2018}. This kind of model promotes the assignment of different weights for the edges at each layer, considering a given node neighborhood (generally first-order neighbors \cite{Velickovic2018}). To do so, an attention mechanism \cite{Vaswani2017} performs a self-attention operation on the nodes to generate attention coefficients (weights for the edges). These coefficients are used to perform a linear combination between the features of the respective nodes, composing the final output features for every node.

In summary, GATs state the relevance of the node $j$'s features to node $i$. GATs have achieved good results regarding graph learning tasks to other problems \cite{Velickovic2018, Boaz2018}. Therefore, we also evaluated GATs in our framework. Using GAT with HiCoRe-ResNet50, we achieved $62.75\%$ of accuracy with the best hyperparameters of the network (grid-search). In this case, they consisted of the learning rate $= 0.005$, dropout $= 0.3$ and number of neurons $= 128$. Meanwhile, our framework (HiCoRe-ResNet50) with unweight graphs, we obtained an accuracy of $63.86\%$, with the learning rate $= 0.005$, dropout $= 0.5$ and number of neurons $= 128$. From now on, once the traditional ResNet50 obtained good overall results under different analyses, we will use it as our baseline architecture to further experiments (instead of VGG19 and VGG16).

\subsection{MIT67 Dataset} \label{subsec:mit}

The MIT67 dataset \cite{Quantoni2009} contains $67$ classes (subclasses) from indoor scenes covering a wide range of $5$ contexts (superclasses), including ``leisure'', ``working place'', ``home'', ``store'' and ``public space'' scene categories. From this dataset, we can explicitly evaluate the granularities of our hierarchical reasoning strategy which considers superclasses and subclasses. Then, besides considering the bounding-boxes in an image as nodes in a graph structure, we also consider each image in a given superclass as nodes in a supergraph.

In our experiments, we filter out some (sub)classes from the original dataset, due to some problems, such as: few samples to compose the training and test sets, annotation errors and missing data. The subclasses disregarded were: ``auditorium'', ``bowling'', ``elevator'', ``jewellery shop'', ``locker room'', ``hospital room'', ``restaurant kitchen'', ``subway'', ``laboratory wet'', ``movie theater'', ``museum'', ``nursery'', ``operating room'', ``waiting room''. For instance, some classes (elevator, locker room, restaurant kitchen) presents only one image. Then, we considered $53$ subclasses, $5$ superclasses, $2,611$ global images and $50,868$ bounding box images. Table \ref{tab:mit67-dataset-superclasses} shows the data distribution considered in our experiments for the MIT67 dataset. It is also important to highlight that the scope of the present work does not aim to treat issues such as a context zero-shot learning.

\begin{table}[h]
\centering
\caption{Superclasses, number of subclasses, number of images for each superclass and number of bounding boxes for each image from the MIT67 dataset.}
\label{tab:mit67-dataset-superclasses}
\small
\begin{tabular}{lccc}
\hline
superclasses & subclasses & images & bounding boxes\\\hline
leisure & $9$ & $228$ & $4,995$\\
working place & $11$ & $477$ & $9,902$\\
home & $13$ & $1,390$ & $25,074$\\
stores & $12$ & $284$ & $6,945$\\
public spaces & $8$ & $232$ & $3,952$\\\hline
\end{tabular}
\end{table}

We compared HiCoRe against the traditional end-to-end architectures, considering each granularity (superclass, subclass and hierarchical). To do so, we used visual features and ResNet50, since they present the best results.

Our framework outperformed the traditional architecture for all granularities. Considering the coarse-grained level (superclass), HiCoRe-ResNet50 reached an accuracy of almost $99.00\%$, while the traditional architecture reached about $52.00\%$. For the other granularities (subclass and hierarchical), HiCoRe-ResNet50 presents gains of $4.79$ and $1.2$ times greater than ResNet50.

Analyzing the performance of the granularities (Table \ref{tab:results-mit67}), the superclass granularity presents higher accuracies, due to it deals with few ($5$) classes (e.g. the subclass granularity considers $53$ classes). Although the hierarchical granularity showed good results, they can be improved using selection strategies to better decide which nodes of the graph will be connected.

\begin{table}[h]
\centering
\caption{Results obtained by HiCoRe considering different (superclass, subclass and hierarchical) granularities.}
\label{tab:results-mit67}
\small
\begin{tabular}{lc}
\hline
 & {HiCoRe-ResNet50}\\\hline
 {visual-superclass} & $99.00\%$\\
 {visual-subclass} & $69.98\%$\\
 {visual-hierarchical} & $58.96\%$\\\hline
\end{tabular}
\end{table}

\subsection{Visual Relationship Detection Dataset} \label{subsec:vrd}

The Visual Relationship Detection (VRD) dataset \cite{Lu2016} is composed of different interactions between pairs of objects. The interactions are verbs (e.g. wear), spatial (e.g. on top of), prepositions (e.g. with), comparative (e.g. taller than), actions (e.g. kick) or a preposition phrase (e.g. drive on).

For our experiments, we need to define the global classes for the bounding-boxes of the images from the dataset. To do so, different heuristics can be applied (e.g. random, first predicate, higher occurrence, center viewing bias). Some heuristics can present problems. For example, the predicate selected in a randomized way can impair the reproducibility of the experiments. The first predicate (related to the first analyzed predicate) and the predicate of higher occurrence can present problems regarding the unbalancing of samples, generating few global classes and with the predominance of samples from the global class \emph{on}.

In this sense, the predicate selection based on the center viewing bias is an interesting alternative. It reflects a tendency to look straight-ahead, to a location which typically coincides with the scene center in visual perception experiments \cite{Bindemann2010}. In photographic images, objects of interest often provide a focal point in a central location that could therefore give rise to a central viewing effect. This effect is also found when observers are searching for people in visual scenes.
  
After the definition of the global classes, we defined the subset of bounding-boxes for the images from the original VRD dataset. For each image, considering the predicate obtained through the adopted heuristic, we can obtain the selection of all the bounding boxes with the same class of the obtained predicate or the $k$ pairs of bounding boxes under this same condition, closest to the center of the image.

In our experiments, we obtained the global classes ($23$ subclasses) with at least $16$ images (Table \ref{tab:vrd-dataset}), in order to ensure samples from all classes for the training and testing sets, as well as for the labeled and unlabeled sets of the semi-supervised approach. To the experiments considering the semi-supervised approach the training set was divided into labeled and unlabeled subsets, comprising $20\%$ and $80\%$, respectively. Hence, we extremely reduced the labeled dataset. To evaluate the results we used accuracy on VRD because our approach focus on the global context of the image, differently from works like \cite{Feifei2017} that use the same dataset.

\begin{table}[h]
\centering
\caption{Global classes (subclasses) obtained through the centering viewing bias, number of images for each subclass and number of bounding boxes for each image from the VRD dataset.}
\label{tab:vrd-dataset}
\begin{tabular}{lcc}
\hline
{global classes} & {images} & {bounding boxes}\\
\hline
above & $204$ & $3,463$\\
against & $16$ & $298$\\
attached to & $21$ & $404$\\
behind & $324$ & $5,951$\\
below & $73$ & $1,152$\\
beside & $63$ & $1,195$\\
by & $44$ & $756$\\
carrying & $17$ & $308$\\
has & $932$ & $17,161$\\
holding & $98$ & $2,001$\\
in & $305$ & $5,925$\\
in front of & $214$ & $3,801$\\
inside & $16$ & $254$\\
left of & $37$ & $737$\\
near & $117$ & $1,932$\\
next to & $230$ & $4,213$\\
on & $1,217$ & $23,607$\\
on top of & $36$ & $678$\\
over & $49$ & $837$\\
right of & $21$ & $422$\\
wearing & $468$ & $9,425$\\
with & $29$ & $550$\\
under & $151$ & $2,626$\\
\hline
\end{tabular}
\end{table}

Table \ref{tab:vrd-results} shows the results comparing the supervised and semi-supervised HiCoRe approaches against the traditional ResNet50 one. From these results it is possible to see that both (supervised and semi-supervised) approaches presented an accuracy gain of approximately $55\%$ in comparison with the traditional architecture. Although, the supervised and semi-supervised approaches practically ties considering the accuracy, the semi-supervised HiCoRe uses a training set $5$ times smaller and reached the same results of the supervised approach.

\begin{table}[h]
\centering
\caption{Results obtained by the (supervised and semi-supervised) HiCoRe approaches on VRD dataset. Symbol \textbf{`--'} refers to the absence of the traditional semi-supervised ResNet50 approach.}
\label{tab:vrd-results}
\begin{tabular}{lcc}
\hline
& {HiCoRe-ResNet50} & {ResNet50}\\\hline
{supervised} & $30.78$ & $19.86$\\
{semi-supervised} & $30.42$ & -- \\
\hline
\end{tabular}
\end{table}

\section{Conclusion and Future Work} 
\label{sec:conclusion}

We have introduced a novel framework for visual Hierarchical Context-Reasoning (HiCoRe). The proposed framework is capable of defining the global context of images under challenging (ambiguous) scenarios. Besides the intrinsic features, to reach this prediction, HiCoRe grasp and describe the semantic relationships among objects from an image (in fine-grained level) and interactions between images from a same context (in a coarse-grained level).

The experiments testify that our framework outperforms traditional state-of-the-art CNN architectures by a considerable margin, reaching a gain $3$ times greater on well-known image datasets. We performed an extensive experimental evaluation to analyze the behavior of HiCoRe under different scenarios.

In future works we intend to apply edge-based feature methods and accelerate the graph convolutions using sampling techniques. In addition, other policies can be used to build the graphs.

\section{Acknowledgements}

This work has been supported by CNPq (grants $\#431668/2016$-$7$, $\#422811/2016$-$5$), CAPES, Arauc\'{a}ria Foundation, SETI, UMD, UMIACS and UTFPR.

{\small
\bibliographystyle{ieee}
\bibliography{manuscript}
}

\end{document}